\documentclass[review]{elsarticle}

\usepackage{lineno,hyperref}
\modulolinenumbers[5]

\usepackage{graphicx} 
\usepackage{CJKutf8}
\usepackage[dvipsnames]{xcolor}

\journal{Journal of Online Social Networks and Media}

\begin{document}
\begin{CJK*}{UTF8}{gbsn}

\begin{frontmatter}

\title{SWSR: A Chinese Dataset and Lexicon for Online Sexism Detection}

\author{Aiqi Jiang\textsuperscript{\rm 1}, Xiaohan Yang\textsuperscript{\rm 2}, Yang Liu\textsuperscript{\rm 1}, Arkaitz Zubiaga\textsuperscript{\rm 1}
  }

\address{\textsuperscript{\rm 1}Mile End Road, London E1 4NS, \textsuperscript{\rm 2}Wheatley Campus, Oxford OX33 1HX}

\author{\textsuperscript{\rm 1}Queen Mary University of London}
\cortext[mycorrespondingauthor]{This is to indicate the corresponding author.}
\ead{a.jiang@qmul.ac.uk}

\author{\textsuperscript{\rm 2}Oxford Brookes University}

\begin{abstract}

Online sexism has become an increasing concern in social media platforms as it has affected the healthy development of the Internet and can have negative effects in society. While research in the sexism detection domain is growing, most of this research focuses on English as the language and on Twitter as the platform. Our objective here is to broaden the scope of this research by considering the Chinese language on Sina Weibo. We propose the first Chinese sexism dataset -- Sina Weibo Sexism Review (SWSR) dataset --, as well as a large Chinese lexicon SexHateLex made of abusive and gender-related terms. We introduce our data collection and annotation process, and provide an exploratory analysis of the dataset characteristics to validate its quality and to show how sexism is manifested in Chinese. The SWSR dataset provides labels at different levels of granularity including (i) sexism or non-sexism, (ii) sexism category and (iii) target type, which can be exploited, among others, for building computational methods to identify and investigate finer-grained gender-related abusive language. We conduct experiments for the three sexism classification tasks making use of state-of-the-art machine learning models. Our results show competitive performance, providing a benchmark for sexism detection in the Chinese language, as well as an error analysis highlighting open challenges needing more research in Chinese NLP. The SWSR dataset and SexHateLex lexicon are publicly available\footnote{http://doi.org/10.5281/zenodo.4773875}.

\end{abstract}

\begin{keyword}
Chinese Sexism Dataset\sep Sexism Detection\sep Hate Speech Detection \sep Abusive Language Detection \sep Chinese Sexist Lexicon \sep Natural Language Processing

\end{keyword}

\end{frontmatter}

\linenumbers

\section{Introduction}

Along with an unprecedented ability for communication and information sharing, social media platforms provide an anonymous environment which allows users to take aggressive attitudes towards specific groups or individuals by posting abusive language \cite{fortuna2021how}. This leads to increasing occurrences of incidents, hostile behaviours and remarks of harassment, especially for online interactions between people of different genders, nationalities, ethnicities, cultures and physical appearances \cite{nobata2016abusive,fersini2018evalita,fersini2020ami,chiril-etal-2020-said,pamungkas2021joint}. Hate speech is one of the most important conceptual categories in anti-oppression politics today \cite{gagliardone2015countering,williams2019hate}, referring to using the language to incite violence or to promote hatred against particular groups of people, or to attack, to insult or to disparage members of a group on the basis of specific characteristics \cite{fortuna2018survey}. Sexism is a common pattern of hate speech and is currently considered as a deteriorating factor in social networks in China \cite{waseem2016hateful,frenda2019online,shi2020perception}. Sex is a sensitive topic in Asian cultures, hence many women still have a high cognitive and tolerance threshold for hostile gender-biased behaviours \cite{shi2020perception}, which consequently aggravates abusive remarks and violent behaviours online. The task of mitigating hate speech online has attracted the attention of Chinese industries, such as Sina Weibo, to impose strict censorship on the contents of relevant topics \cite{deluca2016weibo}, but has remain largely understudied in academic research.

In the past few years, due to the increasing amount of user-generated content and the diversity of user behaviour towards women in social media, manual inspection and moderation of sexist contents becomes unmanageable. The academic community has seen a rapid increase in research tackling the automatic detection of misogynous behaviour and gender-based hatred in both monolingual and multilingual scenarios \cite{jha2017does,rodriguez2020automatic}. The first attempt was by Hewitt et al. \cite{hewitt2016problem} who investigated the manual classification of misogynous tweets, and the first survey of automatic misogyny identification in social media was conducted by Anzovino et al. \cite{anzovino2018automatic}. Nozza et al. \cite{nozza2019unintended} attempted to measure and mitigate unintended bias in machine learning models for misogyny detection. An extensive of misogyny detection is then conducted especially in multilingual and cross-domain scenarios \cite{pamungkas2020misogyny}.

\begin{figure}[!htb]
\centerline{\includegraphics[width=0.9\textwidth]{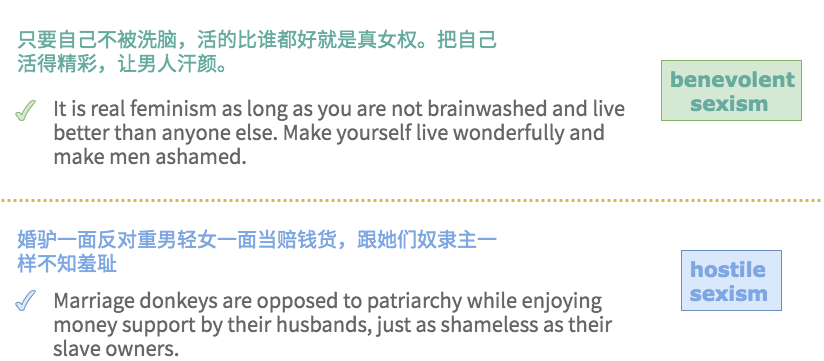}}
\caption{Examples of hostile and benevolent sexism.}
\label{fig:sex.eg}
\end{figure}

However, misogyny is not always equivalent to sexism, and frequently implies the expression of hostility and hatred against women \cite{pamungkas2020misogyny}. As for sexism, Glick and Fiske \cite{glick2001ambivalent} define the concept of sexism referring to two forms of sexism: hostile sexism and benevolent sexism. Hostile sexism is characterised by an explicitly negative attitude towards women, while benevolent sexism is more subtle with seemingly positive characteristics (see examples in Figure \ref{fig:sex.eg}). Sexism includes a wide range of behaviours (such as stereotyping, ideological issues, sexual violence, etc.) \cite{manne2017down,anzovino2018automatic}, and may be expressed in different ways: direct, indirect, descriptive or reported \cite{hellinger200821,chiril-etal-2020-said}. Thus, misogyny is only one case of sexism \cite{manne2017down}. Most previous studies concentrate more on detecting hostile and explicit sexism, overlooking subtle or implicit expressions of sexism \cite{waseem2016hateful,anzovino2018automatic,frenda2019online,pamungkas2020misogyny}. Hence, dealing with the detection of sexism in a wide spectrum of sexist attitudes and behaviours is necessary as these are, in fact, the most frequent and dangerous for society \cite{richardson2018woman}.

Most relevant studies for identifying online abusive content against women utilise supervised approaches, and recently, deep learning approaches have become more popular, especially transformer-based approaches, which have made state-of-the-art achievements in different languages \cite{parikh2019multi,rodriguez2020automatic,chiril-etal-2020-said,samory2021callme}. Since these approaches for automatic sexism detection are usually established utilising labeled training data, the performance is more dependent on the quality and taxonomy of the available datasets \cite{kiritchenko2020confronting}. Some existing studies have made effort to construct sexism-related datasets instead of only collecting explicit misogyny from diverse social platforms in multiple languages (such as English \cite{jha-mamidi-2017-compliment}, Spanish \cite{rodriguez2020automatic} and French \cite{chiril-etal-2020-annotated}), aiming to improve the task performance of detecting online sexist behaviour in a broad sense.
One of the limitations for those approaches is not taking any additional domain knowledge into consideration, like linguistic information from the domain-specific lexicon \cite{wiegand2018inducing}. Recently, several works have demonstrated the positive influence of infusing external domain knowledge on hate speech detection, a broader research field than sexism detection for online abuse\cite{koufakou-etal-2020-hurtbert}, but there is still a lack of more relevant research.


When it comes to sexism-related datasets and resources, however, most efforts have been made for Indo-European languages  \citep{waseem2016hateful,fersini2018evalita,fersini2018ibereval,bhattacharya2020developing,mulki2021let}, while the development of Chinese sexism identification is hindered due to the lack of Chinese annotated resources and Chinese sexism-related lexicons. Moreover, the creation of such resources poses several challenges when it comes to data collection and annotation, especially with the diversity of Chinese dialects and the ambiguity brought about by emerging Internet language.

This paper aims to investigate how diverse behaviors, beliefs and attitudes towards women are expressed in social media, and to focus on collecting data resources about sexism in Chinese. Given the modest presence of Chinese content and geographical access restrictions on Twitter, here we focus on the most prevalent microblogging platform in China, Sina Weibo. As a platform integrating the major features of Twitter, Facebook, and Instagram, users of Sina Weibo can share posts (weibos) with texts, photos, and videos, which can trigger replies between users (comments) and endorsement (likes) from others \cite{sinaweibo,weibo-stats}. In this paper, we make the first effort in creating a sexism dataset in Chinese, to the best of our knowledge. By using Sina Weibo to collect sexism-related weibos and comments, we build, annotate and analyse the Sina Weibo Sexism Review (SWSR) dataset. The SWSR dataset consists of two parts: \textit{SexWeibo} and \textit{SexComment}, both of which include the textual content of posts along with anonymised information of users, number of likes and other metadata. The process led to a dataset with 1,527 weibos and 8,969 comments. In addition, with the aim of assisting research in detection and analysis of sexist comments in Chinese, we provide a sexism-related offensive lexicon SexHateLex which aggregates and extends existing lexical resources in Chinese. Furthermore, we present the first experimentation in Chinese sexism detection to provide a benchmark, including the implementation of various machine learning and deep learning methods. Our experiments and methodology for sexism detection aims to further research in this task in Chinese, as well as enables similar research efforts in other types of hate speech detection in Chinese. Our Chinese dataset and lexicon also enable multilingual sexism research which break the restriction of limited language resources. Abundant demographic and Weibo-based features in SWSR empower to exploit relevant studies on online abusive language in different aspects.

\subsection{Objectives}

To address the problem of the scarcity of Chinese resources in the field of hate speech especially for gender-related content in social media, we focus on the following research objectives:

\begin{itemize}
    \item \textit{Objective 1}: To define a methodology for the collection and annotation of Chinese online sexism at different levels of granularity, involving both explicitly hostile sexism and implicitly subtle sexism. This research builds on and adapts existing annotation guidelines for other languages, providing the first such effort in Chinese in sexism and hate speech.
    
    \item \textit{Objective 2}: To evaluate the effectiveness of existing state-of-the-art models on detecting Chinese sexist content.
    
    \item \textit{Objective 3}: To create a Chinese sexist lexicon to assist research in detection and analysis of Chinese sexist content and assess the influence of external lexical knowledge on the model performance.

\end{itemize}

\subsection{Contribution}

The main novel contributions of this paper are summarised as follows: 

\begin{enumerate}
    \item We construct and release the first Chinese sexism dataset to our knowledge. The rich features of our SWSR dataset including weibo contents, weibo reviews and basic user information make it possible to detect sexist content with various approaches for better performance and interpretability, as well as enables contextual analysis of sexism. 
    \item We further provide labels for the sexism category and the type of target of sexist comments, which enables finer-grained investigation of sexist texts.
    \item We integrate existing lexical resources and sexism-related terms to build a lexicon including 3,016 sexist and abusive terms, which can support research on Chinese abusive language.
    \item We perform an exploratory analysis to validate the quality of the dataset and to understand how sexism is manifested in Chinese.
    \item We present preliminary experiments along with an analysis of the results, establishing a benchmark for Chinese sexism detection.
\end{enumerate}

\subsection{Paper Structure}

This paper is organised as follows. We introduce related work on this task in Section \ref{sec:related}, starting with several previous studies for existing sexism-related datasets. We also present recent work and resources focusing on hate-related lexicons. In Section \ref{sec:collection} we describe the process of collecting and organising source data from Sina Weibo. Section \ref{sec:annotation} presents guidelines and evaluation of three annotation tasks for the collected dataset. The procedure of building a sexist lexicon is introduced in Section \ref{sec:lexicon}. Then we describe experimental results and analysis for sexism detection in Section \ref{sec:experiment}. Section \ref{sec:application} discusses potential areas of research enabled by our dataset and lexical resources. Section \ref{sec:fair} briefly discusses that our work adheres to the `FAIR' facets. Finally, we present conclusive remarks for our work in Section \ref{sec:conclusion}.

\section{Related Work}
\label{sec:related}

Since no previous datasets in Chinese exist for online abusive language targeting gender groups, we discuss recent sexism-related datasets in non-Chinese language in the literature, and review existing lexical resources in Chinese relevant to sexism and online abuse.

\subsection{Existing Non-Chinese Datasets for Sexism Detection}

\begin{table*}[!htb]
\centering
\caption{Existing sexism-related datasets in multiple languages.}
\label{tab:dataset}
\resizebox{0.99\textwidth}{!}{%
\begin{tabular}{lllllll}
\hline
\textbf{Dataset} & \textbf{Language}                                                & \textbf{Offense Type}                                                     & \textbf{Label}                                                             & \textbf{Instance}                                                                               & \textbf{Year} & \textbf{Ref}                                                  \\ \hline
Waseem\&Hovy     & English                                                          & \begin{tabular}[c]{@{}l@{}}Racism \\ Sexism\end{tabular}                  & \begin{tabular}[c]{@{}l@{}}racism, sexism\\ neither\end{tabular}          & 16k                                                                                             & 2016          & \cite{waseem2016hateful}                           \\ \hline
Jha\&Mamidi      & English                                                          & \begin{tabular}[c]{@{}l@{}}Ambivalent \\ sexism\end{tabular}              & \begin{tabular}[c]{@{}l@{}}benevolent\\ hostile, others\end{tabular}      & 22142                                                                                           & 2017          & \cite{jha-mamidi-2017-compliment}                  \\ \hline
AMI@Evalita      & \begin{tabular}[c]{@{}l@{}}English\\ Italian\end{tabular}        & Misogyny                                                                  & \begin{tabular}[c]{@{}l@{}}misogynous\\ not misogynous\end{tabular}        & \begin{tabular}[c]{@{}l@{}}5000(EN)\\ 5000(IT)\end{tabular}                                     & 2018          & \cite{fersini2018evalita}                          \\ \hline
AMI@IberEval     & \begin{tabular}[c]{@{}l@{}}English\\ Spanish\end{tabular}        & Misogyny                                                                  & \begin{tabular}[c]{@{}l@{}}misogynous\\ not misogynous\end{tabular}        & \begin{tabular}[c]{@{}l@{}}3977(EN)\\ 4138(ES)\end{tabular}                                     & 2018          & \cite{fersini2018ibereval}                         \\ \hline
Chowdhury et al. & English                                                          & \begin{tabular}[c]{@{}l@{}}Sexual \\ harassment\end{tabular}              & \begin{tabular}[c]{@{}l@{}}recollection\\ not recollection\end{tabular}    & 5119                                                                                            & 2019          & \cite{chowdhury2019youtoo}                         \\ \hline
HatEval          & \begin{tabular}[c]{@{}l@{}}English\\ Spanish\end{tabular}        & \begin{tabular}[c]{@{}l@{}}Against\\ immigrants \\ and women\end{tabular} & \begin{tabular}[c]{@{}l@{}}non-hateful\\ hateful\end{tabular}              & \begin{tabular}[c]{@{}l@{}}13000(EN)\\ 6600(ES)\\ 9091(immigrants) \\ 10509(women)\end{tabular} & 2019          & \cite{basile2019semeval}                           \\ \hline
Parikh et al.    & English                                                          &  Sexism            & 23 categories of sexism                                                               & 13k                                                                                             & 2019          & \cite{parikh2019multi}                             \\ \hline
TRAC-2           & \begin{tabular}[c]{@{}l@{}}English\\ Hindi\\ Bangla\end{tabular} & \begin{tabular}[c]{@{}l@{}}Misogyny \\ Aggression\end{tabular}            & \begin{tabular}[c]{@{}l@{}}GEN(Gendered)\\ NGEN(Non-gendered)\end{tabular} & 25k                                                                                             & 2020          & \cite{bhattacharya2020developing}                  \\ \hline
AMI@Evalita      & Italian                                                          & Misogyny                                                                  & \begin{tabular}[c]{@{}l@{}}misogynous\\ not misogynous\end{tabular}        & 7000                                                                                            & 2020          & \cite{fersini2020ami}                              \\ \hline
Chiril et al.      & French                                                          & Sexism                                                                & \begin{tabular}[c]{@{}l@{}}direct, descriptive\\ reporting, non-sexist \\ no decision\end{tabular}        & 12k                                                                                            & 2020          & \cite{chiril-etal-2020-annotated}                              \\ \hline
MeTwo            & Spanish                                                          & Sexism                                                                    & \begin{tabular}[c]{@{}l@{}}sexist, not-sexist\\ doubtful\end{tabular}               & 3600                                                                                            & 2020          & \cite{rodriguez2020automatic}                      \\ \hline
EXIST@IberLEF    & \begin{tabular}[c]{@{}l@{}}English \\ Spanish\end{tabular}       & Sexism                                                                    & \begin{tabular}[c]{@{}l@{}}sexist\\ not sexist\end{tabular}                & \begin{tabular}[c]{@{}l@{}}5644(EN) \\ 5741(ES)\end{tabular}                                    & 2021          & -           \\ \hline
LeT-Mi           & Arabic                                                           & Misogyny                                                                  & \begin{tabular}[c]{@{}l@{}}misogynistic\\ non-misogynistic\end{tabular}    & 6550                                                                                            & 2021          & \cite{mulki2021let}                                \\ \hline
Guest et al.      & English                                                          & Misogyny                                                                  & \begin{tabular}[c]{@{}l@{}}misogynistic\\ non-misogynistic\end{tabular}    & 6567                                                                                            & 2021          & \cite{guest2021expert}                             \\ \hline
ArMI@HASOC       & Arabic                                                           & Misogyny                                                                  & \begin{tabular}[c]{@{}l@{}}misogynistic\\ non-misogynistic\end{tabular}    & 9833                                                                                            & 2021          & - \\ \hline
Samory et al.    & English                                                          & Sexism                                                                    & \begin{tabular}[c]{@{}l@{}}benevolent\\ hostile, other\\ callme, scale\end{tabular} & 16k                                                                                             & 2021          & \cite{samory2021callme}                            \\ \hline
Bajer            & Danish                                                           & Misogyny                                                                  & \begin{tabular}[c]{@{}l@{}}misogyny\\ racism, others\end{tabular}                    & 27.9k                                                                                           & 2021          & \cite{zeinert2021annotating} \\ \hline
\end{tabular}%
}
\end{table*}

The last few years have witnessed an increase in the interest in and availability of sexism datasets. We provide a summary of the existing sexism datasets in Table \ref{tab:dataset}. The earliest attempt was that by Waseem and Hovy \cite{waseem2016hateful}, who provided a publicly available dataset of more than 16k tweets for hate speech and annotate it into three categories - racism, sexism and neither. However, it only comprises the expression of hostile sexism towards women, overlooking other kinds of sexism. Chowdhury et al. \cite{chowdhury2019youtoo} aggregate experiences of sexual abuse to facilitate a better understanding of social media construction and to bring about social change. These two datasets consist of content in English.

In addition, recent sexism datasets include multilingual content involving Italian, Spanish and Hindi, along with English. The Automatic Misogyny Identification (AMI) competitions in Evalita 2018 \cite{fersini2018evalita}, Ibereval 2018 \cite{fersini2018ibereval} and Evalita 2020 \cite{fersini2020ami} provide datasets in English, Spanish and Italian to detect misogynistic content, to classify misogynous behaviour as well as to identify the target of a misogynous text. HatEval@SemEval 2019 \cite{basile2019semeval} is another competition aiming to detect hate speech against immigrants and women and further finer-grained features in offensive text, like aggressive attitude and the target harassed in English and Spanish posts from Twitter. Furthermore, Parikh et al. introduce a dataset consisting of accounts of sexism in 23 categories to investigate sexism categorisation as a multi-label classification task \cite{parikh2019multi}. Bhattacharya et al. develope a multilingual annotated corpus of misogyny and aggression in Indian English, Hindi, and Indian Bangla as part of a project studying and automatically identifying misogyny and communalism in social media \cite{bhattacharya2020developing}. The first French dataset \cite{chiril-etal-2020-annotated} and Spanish dataset (MeTwo) \cite{rodriguez2020automatic} have been released for sexism detection, and EXIST@IberLEF 2021\footnote{http://nlp.uned.es/exist2021} proposes the first shared task on sexism identification in social networks (as opposed to misogyny detection), aiming to detect online sexism in English and Spanish. Moreover, Hala and Bilal \cite{mulki2021let} introduce the first Arabic Levantine dataset for online Misogyny (LeT-Mi) written in the Arabic and Levantine dialect. Then ArMI@HASOC 2021 at FIRE\footnote{https://sites.google.com/view/armi2021/} proposes an Arabic Misogyny Identification (ArMI) task with two sub-tasks derived from the Let-Mi dataset\cite{mulki2021let}, which is the first shared task to address the problem of automatic detection of Arabic online misogyny. Guest et al. \cite{guest2021expert} introduce an expert annotated misogynous dataset collected from Reddit and present a new detailed hierarchical taxonomy for online misogyny, while Zeinert et al. develop the first Danish misogyny dataset, Bajer, under a four-level taxonomy of labels \cite{zeinert2021annotating}. Besides, Samory et al. provide a sexism dataset using psychological scales and generating adversarial samples to improve construct validity and reliability in sexism detection \cite{samory2021callme}.

Despite the increasing availability of sexism datasets, in an increasing number of Indo-European languages, no dataset exists in the Chinese language \cite{vidgen2020directions}. Likewise, we are not aware of previous research in sexism detection in Chinese. To fill this gap, our research here documents our efforts in creating the first such dataset, including Chinese social media posts labelled as sexist or not. For sexist posts, we annotate the sexism category and target type as well to support deeper investigation on sexism identification.

\subsection{Existing Lexical Resources for Online Abuse}

Detection of offensive content can be challenging as it not always contains explicit mentions of negative or hateful words \cite{han2020fortifying,fortuna2021how}. However, there is evidence showing that the use of domain-specific lexical words in classification models can boost model performance \cite{schmidt2017survey,mladenovic2017using,koufakou-etal-2020-hurtbert}. With the expectation that the use of a lexicon can make for a good proxy to improve detection of hate speech, here we develop one in Chinese to support our research in sexism detection. There are many popular lexicons for online abuse, which collect and organise offensive words and phrases. For example, \cite{burnap2016us} focus on several lists obtained from Wikipedia that are particularly linked to a specific sub-type of hate speech in English, such as ethnic slurs\footnote{https://en.wikipedia.org/wiki/List\_of\_ethnic\_slurs} and LGBT slang terms\footnote{https://en.wikipedia.org/wiki/List\_of\_LGBT\_slang\_terms}. A popular hate speech lexicon is HateBase,\footnote{https://hatebase.org/} which provides the largest multilingual hate speech lexicon linked to aspects such as religion, gender and ethnicity. It includes 3,635 groups of terms in more than 95 languages \cite{tuckwood2017hatebase}. Despite its volume for languages like English, the HateBase lexicon only contains 39 Chinese terms, which is still far from becoming a referential resource. Besides,  \cite{bassignana2018hurtlex} built a multilingual hate speech lexicon, HurtLex\footnote{https://github.com/valeriobasile/hurtlex}, involving over 50 languages. HurtLex provides a larger set of 4,251 terms in Chinese. 

However, there is no relevant study for the Chinese sexism scenario, and only a few Chinese lexical resources are designed for offensive language. None of those resources focuses specifically on gender-related contents. Given the scarcity of sexism-specific lexicons as well as the strong relation between those phenomena of offensive language and sexist language \cite{pamungkas2020misogyny}, we aggregate and expand existing Chinese lexicons to build a large Chinese lexicon consisting of terms that can be generally associated with abusive language as well as gender-specific terms, which can assist by furthering research in Chinese sexism detection.


\section{Data Collection}
\label{sec:collection}

In describing our data collection process, we first describe the key characteristics of the Sina Weibo microblogging platform we use to build our SWSR dataset, discussing the different data harvesting options across the different weibo platforms. Then we delve into the data collection and filtering process.

\subsection{Sina Weibo}

Sina Weibo is the largest microblogging service in China, which has some unique characteristics with respect to Twitter. It is aimed for information sharing, dissemination and information acquisition based on user relationships \citep{sinaweibo}. Content on Sina Weibo is spread through the ``following-follower'' networks established between people \cite{huberman2008social}, for example, allowing users to post comments on someone's Weibo or to reply to other people's comments on someone's Weibo. It allows users to insert images, videos, music, long articles and polls. 

\begin{figure}[!htb]
\centerline{\includegraphics[width=0.9\textwidth]{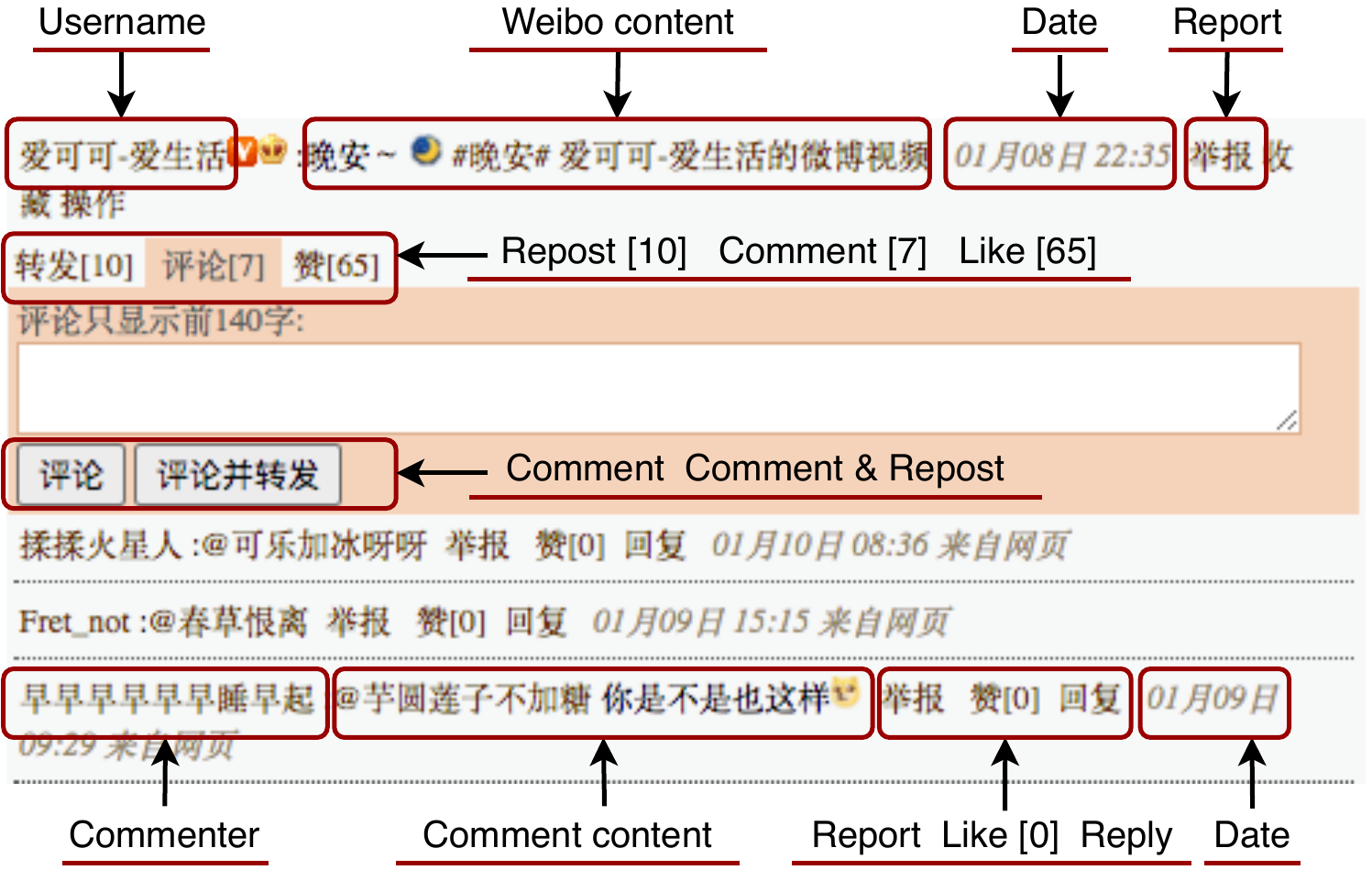}}
\caption{An example of Sina Weibo on weibo.cn}
\label{fig:pccn}
\end{figure}

Sina Weibo has three main ways of accessing its website, namely weibo.com, weibo.cn and m.weibo.com. We can access Sina Weibo via PC terminal through weibo.com and weibo.cn, and the mobile counterpart is m.weibo.com. The weibo.com is more complex than weibo.cn because its Weibo page presents a richer functionality with more components which weibo.cn doesn't have, such as Top Topic Ranking, Hot Movie Recommendation, advertisements, etc. However, we can see in an example of weibo.cn in Figure \ref{fig:pccn} that the website structure is simple and straightforward. Both the weibo and its associated comment list can be easily retrieved and parsed for data collection. So we finally decide to use \textbf{weibo.cn} as the source website of Sina Weibo.

\subsection{Data Collection and Processing}

\begin{figure}[!htb]
\centerline{\includegraphics[width=0.65\textwidth]{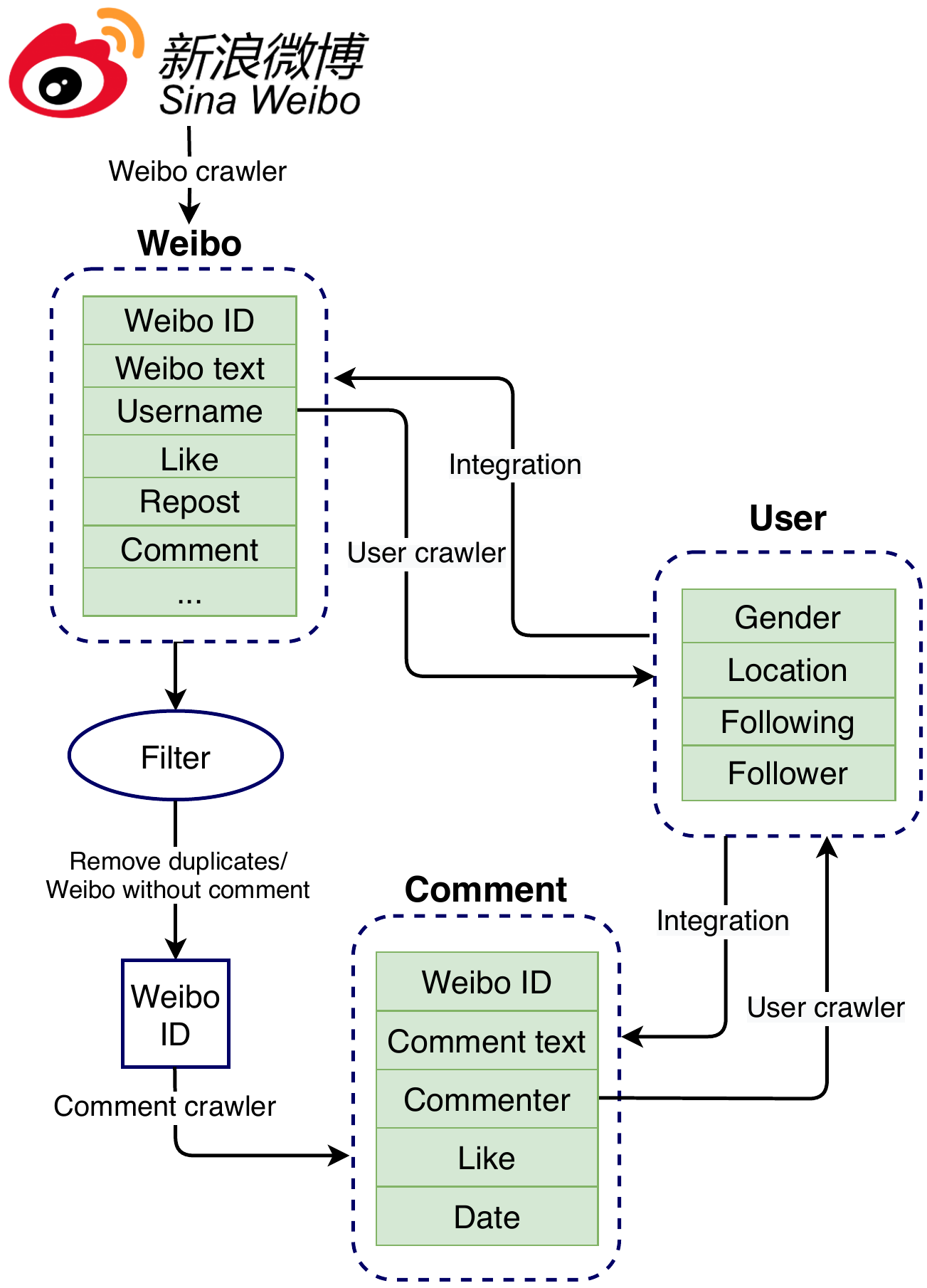}}
\caption{Overview of the data collection process.}
\label{fig:collection}
\end{figure}

As described above, a Sina Weibo timeline comprises posts (weibos) which receive replies (comments). Initially, we use keyword-driven method to collect a set of weibos, for which we then collect the associated comments. While the collection of weibos is restricted to those containing the keywords, our focus on the associated comments allows us more flexibility, retrieving content which need not contain the seed keywords. Figure \ref{fig:collection} shows an overview of the data collection process, which we introduce further details in the steps below. Our SWSR dataset therefore is made of two tables for weibo and comment data along with some anonymised user information pertaining to the weibos and comments. This user information includes features such as user gender and user location. All personally identifiable information is removed and not disclosed, including user names and mentions.

\subsubsection{Step I: Extract Weibo Data}

To construct our dataset, we use keyword-driven search to collect gender-related weibos from Sina Weibo platform (weibo.cn). In terms of relevance to the topic and through manual exploration \cite{richardson2018woman,fersini2018ibereval}, we firstly determine to use seven different keywords related to some hot topics and events of sexism for weibo data collection, namely 婊子 (bitch), 女同性恋 (lesbian), 女权 (feminism), 厌女 (misogyny), metoo运动 (metoo movement), 性别歧视 (gender discrimination) and 性骚扰 (sexual harassment). Then we search and extract weibos containing these keywords. In addition, we retrieve user profiles, which include self-reported values such as gender and location, and other variables such as number of followers. To protect user privacy in the dataset, usernames are anonymised by replacing them with a special token $<$username$>$. Then we combine these features into the weibo. The number of weibos collected for each keyword is listed in Table \ref{tab:keyword}, which amounted to a total of 9,087 weibos collected for all keywords. Data collection was limited to posts made between June 2015 to June 2020.

\begin{table}[!htb]
\centering
\caption{Number of weibos collected for each keyword.}
\label{tab:keyword}
\resizebox{0.8\textwidth}{!}{%
\begin{tabular}{llll}
\hline
\textbf{Keyword} & \textbf{Translation}      & \textbf{Number of Weibos} & \textbf{Total}                 \\ \hline
婊子               & bitch                 & 407                      &  \\ \cline{1-3}
女同性恋             & lesbian               & 520                     &                                \\ \cline{1-3}
女权               & feminism              & 2255                     &                                \\ \cline{1-3}
厌女               & misogyny              & 1757                     &              \textbf{9087}                 \\ \cline{1-3}
metoo运动            & metoo movement        & 1340                     &                                \\ \cline{1-3}
性别歧视             & gender discrimination & 1366                     &                                \\ \cline{1-3}
性骚扰              & sexual harassment     & 1442                     &                                \\ \hline
\end{tabular}%
}
\end{table}

\subsubsection{Step II: Process Weibo Data}

In this step, we process the collected weibos prior to collecting the associated comments in subsequent steps. We remove the weibos that match at least one of the following criteria:

\begin{itemize}
    \item weibos without any comments. This can be easily done by checking the number of comments for each weibo according to `weibo\_comment' column.
    \item duplicates which are exact matches of both the `weibo\_id' and `weibo\_text' columns, i.e. weibos collected repeatedly across keywords. We only keep one of these repeated instances.
\end{itemize}


This led to a final set of 3,856 weibos, along with their associated weibo IDs which we use in the next step to retrieve comments.

\subsubsection{Step III: Extract Comment Data}

In order to extract comments for the collected weibos, we utilise their weibo ID. This enabled us collection of textual content and metadata of weibos, including user profiles of commenters. This led to the collection of 31,677 comments for the 3,856 weibos.

\subsubsection{Step IV: Process Comment Data}

For processing the comments collected in the previous step, we remove comments matching at least one of the following criteria:

\begin{itemize}
    \item Remove duplicate comment texts, keeping only one instance. This is caused by users who copy and paste the same comment repeatedly.
    \item Remove short comments with commonly identified patterns -- fixed tokens on Sina Weibo, e.g. comments solely containing the word `转发' (repost), `回复' (reply) or `举报' (report).
    \item Remove the remaining short comments (length of less than 5 characters).
    \item Remove comments without any Chinese character.
\end{itemize}


Given that users occasionally reply by splitting their texts into multiple comments, we aggregate them. When we find multiple comments from the same user in close temporal proximity, we automatically aggregate them into a single comment.

Finally, we convert all the comments from traditional Chinese to simplified Chinese, which helps ensuring consistency while keeping the same information. We use the Python package chinese\_converter\footnote{\url{https://pypi.org/project/chinese-converter/}} to achieve this.

This led to a final set of 8,969 comments linked to 1,527 weibos, whose statistics are shown in Table \ref{tab:stats}. The final aim of our sexist data collection lies in the retrieval of these comments, which are the ones that we annotate and make up the final dataset. The weibos are solely considered to support the annotation process and, if desired, for context-based analysis of comments.

\subsection{Ethics of Data Collection}

As our dataset is obtained through web crawler built in programming language Python from Sina Weibo platform, we carefully consider the ethical implications behind the collected data. Posts and comments collected in this dataset are in the public domain and web scraping has been done only for research purposes. Hence, we ensure that no ethics approval is needed for this study \cite{xu2020characterizing} and the collected dataset follows acceptable ethical practices by adhering to the following:

\begin{itemize}
    \item Our dataset does not present any personally identifiable information, as we have anonymised all user names in the dataset, including any user names mentioned in the posts (replaced by the special token $<$username$>$)
    \item Our dataset does not include any private messages between users, and there was no interaction between Weibo users and researchers.
    \item We rely on publicly available data and carefully collect the data into multiple steps to avoid overloading Sina Weibo servers.  
    \item The Sina Weibo server is publicly accessible.
\end{itemize}

\section{Data Annotation}
\label{sec:annotation}

During the annotation process of our SWSR dataset, we perform three annotation tasks as follows:

\begin{enumerate}
 \item \textbf{Sexism Identification:} whether a text is sexist, as a binary annotation task determining if a comment is sexist (1) or non-sexist (0). Where a comment is deemed sexist, we also perform two additional annotations:
 \item \textbf{Sexism Category:} We define four categories of sexism, namely stereotype based on appearance (SA), stereotype based on cultural background (SCB), microaggression (MA) and sexual offense (SO).
 \item \textbf{Target Type:} individual (I) or generic (G).
\end{enumerate}

\subsection{Annotation Preparation}

In order to reliably identify sexism as well as their corresponding categories and targets, we provide initial annotation guidelines for all three tasks. The annotation guidelines for sexism identification are based on \cite{waseem2016hateful,chowdhury2019youtoo}, and guidelines for the sexism category and the target type are adapted from \cite{fersini2018ibereval,fersini2018evalita,mulki2021let,richardson2018woman}. Guidelines were iteratively developed through collective annotation of a small sample of 100 comments by a broader set of five annotators. These annotators met and discussed disagreements between them, which led to revised guidelines.

In most cases, we find that our disagreement for annotation task I was mainly caused by the lack of sufficient context when identifying sexist content. For example, one annotator marks the text 它们的大脑平滑到可以在上面溜冰，真的不是一个物种啊 (Their brains are so smooth that they can skate on them. We are really not the same species) as not sexism because there is no sexist content towards women. But when we check the original Weibo text, we find that 'they' in this text means some stupid women who insult men for more benefits. So it should be marked as sexism with consideration of the context. Another common case of disagreement is the misunderstanding of specific words related to sexism. These words commonly appear in sexist text but are not common in general speech. Some annotators did not realise that 婚驴 (marriage donkey) is an offensive word specifically towards women. This word means those women who are as stupid as donkeys in marriage, deprived of a lot of benefits, but still enjoy silly happiness. Discussions following these agreements led to revisions in the guidelines and improvements in subsequent rounds of annotations. In addition, for the annotation task II determining the sexism category, there were disagreements caused by occasional overlaps in the interpretations of the different labels, which were resolved and led to revision of the guidelines. Annotation III consisting in determining the target type was more straightforward as being easier to label.

In what follows, we reproduce initial guidelines used for the three annotation tasks, which enable annotators to have a better understanding of sexist issues for three annotation tasks and to a large extent improve the final score of inner-annotator agreement. 

\subsection{Annotation Guidelines}

Given the difficulty of identifying sexist behaviours, we carefully crafted guidelines for the three annotations tasks based on the insights from the above annotation testing: sexism identification, sexism category and target category, along with examples of annotations by sexism category and target category shown in Table \ref{tab:egs}.

\subsubsection{Annotation I: Sexism Identification}

A comment is considered sexist if it belongs to at least one of the following categories:

\begin{itemize}
    \item explicitly attacks or insults gender groups or individuals using sexist language.
    \item incites gender-based violence or promote sexist hatred but not directly use a sexual abusive language.
    \item abuses those who attack or have negative attitudes towards a gender group.
    \item shows support of problematic incidents or intentions of sexual assault, sexual orientation and sexual harassment.
    \item negatively stereotypes gender groups by describing physical appeal, oversimplifying image or expressing superiority of men over women.
    \item expresses underlying gender bias in a sarcastic or tacit way.
\end{itemize}

The rest of the texts are considered non-sexist. This includes neutral descriptions or testimonies of sex-related events or phenomena.

\subsubsection{Annotation II: Sexism Category}

Each of the comments marked as sexist in the first task needs to be classified into one of the following, determining the sexism category of the comment:

\begin{itemize}
    \item \textit{Stereotype based on Appearance (SA):} describes physical appeal, oversimplifies image, or makes comparison with narrow/vulgar standards towards a gender group.
    
    \item \textit{Stereotype based on Cultural Background (SCB):} expresses opinions indicating the superiority of men over women and emphasises gender inequality under the concept of a patriarchal society.
    
    \item \textit{Microaggression (MA):} intentionally or unintentionally expresses hostile, derogatory or negative attitudes or remarks against gender groups or individuals.
    
    \item \textit{Sexual Offense (SO):} incites sexual-related behaviour or attitude against women, such as sexual harassment, sexual assault, rape and violence.
\end{itemize}

\subsubsection{Annotation III: Target Category}

Each of the comments marked as sexist in the first task needs to have the type of target identified, which can be one of the following two:

\begin{itemize}
    \item \textit{Individual (I):} a post with sexist content addressing a specific person.
    \item \textit{Generic (G):} a post with sexist content addressing a broader group (such as a gender-based group of people).
\end{itemize}

\begin{table}[]
\centering
\caption{Examples of sexism categories and target types in the dataset.}
\label{tab:egs}
\resizebox{\textwidth}{!}{%
\begin{tabular}{llll}
\hline
\textbf{Example}                                                             & \textbf{Translation}                                                                                                                                                                    & \textbf{Sexism Category}                                                          & \textbf{Target} \\ \hline
\begin{tabular}[c]{@{}l@{}}前任的漂亮更清纯甜美\\ 一看就是正经人，现在\\ 这位一看就很肉的感觉\end{tabular} & \begin{tabular}[c]{@{}l@{}}His ex looks more innocent and \\ beautiful, like a decent person. \\ But the appearance of his current \\ girlfriend makes me a higher libido.\end{tabular} &  SA          & I      \\ \hline
\begin{tabular}[c]{@{}l@{}}还是让女性做些带孩子，\\ 鼓励丈夫的工作！\end{tabular}               & \begin{tabular}[c]{@{}l@{}}We should let women do more \\ housework, and encourage their \\ husbands' work!\end{tabular}                                                                & SCB & G          \\ \hline
\begin{tabular}[c]{@{}l@{}}关键是有些女生还没子\\ 宫道德，结了婚脑子里\\ 自动长了个屌\end{tabular}     & \begin{tabular}[c]{@{}l@{}}The point is that some girls have \\ no uterine morals. There is a dick \\ in their head after they get married.\end{tabular}                                & MA                                                                   & G           \\ \hline
\begin{tabular}[c]{@{}l@{}}你全家女性送来给我搞\\ 一搞，我戴套，保证安全\end{tabular}             & \begin{tabular}[c]{@{}l@{}}Send your family’s women to me \\ to fuck them, I will wear a condom \\ to ensure safety\end{tabular}                                                        & SO                                                                    & I     \\ \hline
\end{tabular}%
}
\end{table}

\subsection{Annotator Agreement}

All three annotations were performed independently by three annotators, all of them PhD students, including two females and one male. We use the open source text annotation tool \textit{doccano}\footnote{https://github.com/doccano/doccano} to facilitate the annotation work and to enable independent annotation effectively by three annotators.

We report inter-annotator agreement rates for the three annotators by using Cohen's kappa as a metric \cite{cohen1968weighted}. The inter-annotator agreement of our annotation task I is overall 82.3\% (71.8\% for the sexist class and 96.1\% for non-sexist). For annotation tasks II and III, the inner-annotator agreements reach 76.8\% and 85.5\% respectively. All these agreement rates can be deemed substantial agreements between the three annotators. Examples of annotations by sexism category and target category are shown in Table \ref{tab:egs}.

\section{Lexicon Collection}
\label{sec:lexicon}

We build a large sexism lexicon SexHateLex by aggregating and expanding existing resources, which is a combination of 

\begin{itemize}
    \item profane words and slang,
    \item sexual abusive words and slang, and
    \item sexism-related people, websites and events.
\end{itemize}

SexHateLex is built by integrating four existing lexicons, and augmented by adding typos and synonyms based on integrated sexual-related abusive terms. We aggregate the following lexical resources:

\begin{itemize}
    \item \textit{Chinese Profanity in Wikipedia\footnote{\url{https://en.wikipedia.org/wiki/Mandarin_Chinese_profanity\#Sex}}:} Wikipedia provides a list of Chinese profane words linked to sex, race and sexual orientation. For our purposes we chose the 599 terms for sex and sexual orientation.
    \item \textit{HateBase\footnote{https://hatebase.org/}:} HateBase is the world’s largest structured repository of regionalised multilingual hate speech corpora in the field of religion, gender, nationality, ethnicity, etc. We collect 29 Chinese terms from HateBase.
    \item \textit{TOCP dataset\footnote{http://nlp.cse.ntou.edu.tw/resources/TOCP/}:} TOCP is the largest Chinese profanity dataset including 16,450 sentences \cite{yang2020tocp}. All profane words and corresponding locations in each sentence have been labelled in this dataset. A total of 1,014 profane words are extracted.
    \item \textit{Sexy Lexicon\footnote{https://github.com/fighting41love/funNLP/tree/master/data}:} The repository funNLP provides massive resources to support research in Chinese NLP, one of which is a sexy lexical list in the category of sensitive term datasets. We collect 1,240 terms from this list.
\end{itemize}

After integrating terms from all the resources above, we get a total of 2,109 terms. Then we combine typo words that users make spell mistakes in the text based on a spell checking method in the 'aion' python package\footnote{https://github.com/makcedward/nlp/tree/master/aion}, and add the top 5 similar words to each word in the collected lexical list. Word embeddings\footnote{https://fasttext.cc/} are leveraged for this step, followed by cleaning all duplicate and incorrect terms. This leads to the final SexHateLex lexicon with 3,016 terms.

\section{Data Description}
\label{sec:dataset}

We describe the resulting dataset by first presenting the dataset structure and by then providing descriptive statistics of the dataset.

\subsection{Dataset Structure}

\begin{table}[htb]
\centering
\caption{Description of features in the weibo and comment datasets.}
\label{tab:feature}
\resizebox{0.8\textwidth}{!}{%
\begin{tabular}{ll}
\hline
\textbf{Table} & \textbf{Feature}                                                                                                                                                                                      \\ \hline
SexWeibo      & \begin{tabular}[c]{@{}l@{}}weibo\_id, weibo\_text, keyword, user\_gender, \\ user\_location, user\_follower, user\_following, \\ weibo\_like, weibo\_repost, weibo\_comment, weibo\_date\end{tabular} \\ \hline
SexComment    & \begin{tabular}[c]{@{}l@{}}weibo\_id, comment\_text, gender, location, like, \\  date, label, category, target\end{tabular}                                                                                \\ \hline
\end{tabular}%
}
\end{table}

The SWSR dataset is organised in two files: \textit{SexWeibo.csv} (SexWeibo) and \textit{SexComment.csv} (SexComment), containing weibos (posts) and comments (replies) respectively. Contents in these two files can be linked through the \textit{weibo\_id}. We list all features in \textit{SexWeibo.csv} and \textit{SexComment.csv} files in Table \ref{tab:feature} (see more details in \ref{sec:appendix-structure}). Considering the user privacy, all user names in this dataset are anonymous with a special token $<$username$>$.

\subsection{Dataset Statistics}

\begin{table}[htb]
\centering
\caption{Statistics of the dataset.}
\label{tab:stats}
\resizebox{0.7\textwidth}{!}{%
\begin{tabular}{lccc}
\hline
 & \textbf{All}       & \textbf{Sexist}                   & \textbf{Non-Sexist}    \\ \hline
All                                                                    & 8969                                                    & 3093 (34.5\%) & 5876 (65.5\%) \\ \hline
\begin{tabular}[c]{@{}l@{}}Average length \\ per comment\end{tabular}  & 71.45                                                   & 90.34                                                  & 61.51                                                   \\ \hline
\begin{tabular}[c]{@{}l@{}}Number of comment \\ per weibo\end{tabular} & 5.87                                                    & 3.77                                                   & 4.69                                                    \\ \hline
\end{tabular}%
}
\end{table}

\begin{figure}[!htb]
\centerline{\includegraphics[width=0.65\textwidth]{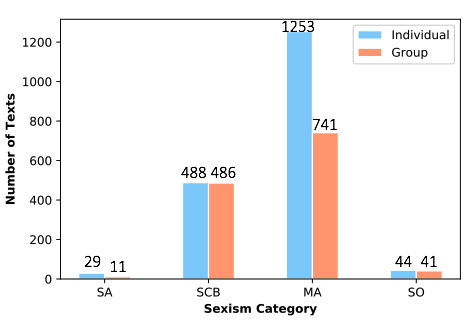}}
\caption{Distribution of sexism categories and target types in the dataset.}
\label{fig:dis}
\end{figure}

\begin{figure}[!htb]
\centerline{\includegraphics[width=0.55\textwidth]{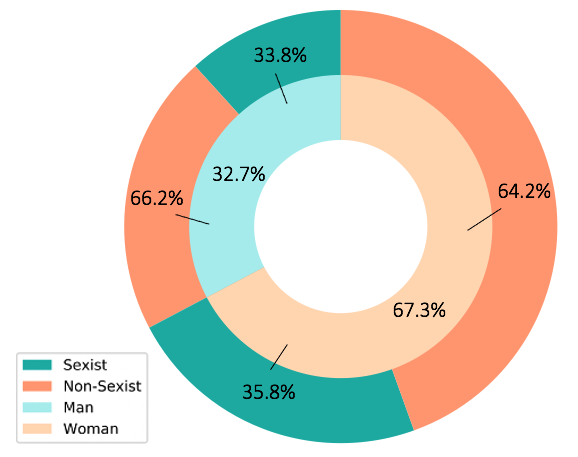}}
\caption{Distribution of user gender across two classes in the dataset.}
\label{fig:gender}
\end{figure}

The resulting 8,969 comments are associated with 1,527 weibos. Table \ref{tab:stats} shows the statistics of the dataset in terms of the distribution of sexist comments, comment length and number of comments per weibo. We can see that the majority of comments are non-sexist, with nearly twice as many as sexist comments.

Figure \ref{fig:dis} depicts the distribution of the sexism category and target type in sexist comments. More than half of the sexist comments are MA, and SCB also takes a large proportion in the sexist class. Besides, the number of comments towards individuals nearly double those towards groups, where sexist texts in the MA category are more frequently abusive towards individuals.

\subsubsection{Textual Distribution}

We compute the average lengths (in a number of characters) of comments in each category. We see big differences in Table \ref{tab:stats} showing that the average length of a sexist comment is 50\% bigger than the length of a non-sexist comment. 
Furthermore, Table \ref{tab:stats} presents the averaged number of comments for each weibo. We can see that the number of comments per weibo for both sexist and non-sexist classes are less than that for all data, which is because that one weibo might contain multiple comments in different classes. Hence, the sum of weibo counts for two classes can be larger than the overall number of weibos.

\subsubsection{Gender Distribution}

According to Figure \ref{fig:gender}, we find that the gender distribution in sexism is skewed towards women while the probabilities of men and women to send sexist posts are similar. As some weibos are more relevant to topic keywords like misogyny and feminism, female users are more likely to attack or insult males who make malicious remarks or show unequal attitudes towards women.

\begin{table*}[htb]
\centering
\caption{Description of the 12 most frequent terms in the dataset (DataTerm) and in the lexicon (LexTerm). [尸吊] is a sensitive character which cannot be found in the Latex package. The table presents the character by dividing it into two parts, which can be easily understood in Chinese. PCT denotes the percentage of each term.}
\label{tab:words}
\resizebox{0.95\textwidth}{!}{%
\begin{tabular}{cccccc}
\hline
\textbf{DataTerm} & \textbf{Translation} & \textbf{PCT} & \textbf{LexTerm} & \textbf{Translation}        & \textbf{PCT} \\ \hline
女权                    & feminism         & 29.84\%             & 骂                     & curse                   & 7.45\%              \\
女性                    & women            & 25.20\%             & 死                     & die                     & 2.89\%              \\
不是                    & not              & 19.04\%             & 搞                     & flirt                   & 2.75\%              \\
男人                    & man              & 11.92\%             & 女拳                    & negative feminism       & 2.20\%              \\
孩子                    & children         & 8.78\%              & 歧视                    & discrimination          & 2.04\%              \\
骂                     & curse            & 7.45\%              & 驴                     & donkey                  & 1.88\%              \\
男权                    & patriarchal      & 6.14\%              & [尸吊]                    & dick                    & 1.78\%              \\
极端                    & extreme          & 5.90\%              & 逼                     & pussy                   & 1.45\%              \\
结婚                    & marry            & 5.26\%              & 强奸                    & rape                    & 1.44\%              \\
姓                     & surname          & 5.26\%              & 狗                     & dog(similar use as pig) & 1.23\%              \\
权利                    & right            & 3.89\%              & 干                     & fuck                    & 1.08\%              \\
平等                    & equality         & 3.73\%              & 蛆                     & maggot                  & 0.89\%      \\ \hline       
\end{tabular}%

}
\end{table*}

\subsubsection{Word Frequency Distribution}

We normalise the data by removing stop words, special markers such as '转发' (Repost), user names, and punctuation marks. Then we select the list of 12 words with the highest frequency in the comments as well as the top 12 words from the SexHateLex lexicon which are most frequent in the comments. We find that the terms frequently occurring in each class differ significantly (see Table \ref{tab:words}). The most frequent tokens in the lexicon present negative emotional attitudes while those in the comments are mostly neutral words related to gender topics.

\section{Preliminary Experiments: Sexism Detection}
\label{sec:experiment}

To assess the difficulty of computationally detecting sexist comments in SWSR and to provide benchmark experimental results, we conduct both coarse-grained and fine-grained sexism detection experiments, evaluating different features and models. Our experiments are designed in three steps:

\begin{enumerate}
    \item Sexism identification (Binary): weibo contents are classified as either sexist or non-sexist. 
    \item Sexism category classification (Multi-class): texts are classified into one of five categories: stereo-type based on appearance (SA), stereotype based on cultural background(SCB), microaggression (MA), sexual offense (SO), or non-sexist.
    \item Target classification (Multi-class): texts are classified into either generic, individual, or non-sexist.

\end{enumerate}

\subsection{Models}

For the three experimental steps, we test various models. As context-based models we utilise different BERT-based models \cite{devlin2019bert} based on transformers. We use three different BERT-based models: (1) BERT, (2) BERT with whole word mask (Bert-wwm), and (3) RoBERTa \cite{liu2019roberta}. Besides, we adopt three different baselines using combinations of unigrams to trigrams as features: (1) a logistic regression (LR), (2) a support vector machine (SVM), and (3) a character-level LR. We also test two content-based models, a CNN and a character-level CNN \cite{kim2014convolutional} with FastText word embeddings. 

In addition, for the experimental step 1, we test all the models above with and without lexical words from the SexHateLex lexicon, to show its impact on the task. We first count the occurrence of each word, and then convert the count vector from the count frequency to term frequency–inverse document frequency (TF-IDF) \cite{jing2002improved}, indicating how significant a category is to a text in the corpus. Finally, we concatenate the TF-IDF lexical vector with textual embeddings.

\subsection{Experiment Settings}

Given that the SWSR dataset is not balanced, especially in the category classification task, we randomly split the comment data into 90\% for training and 10\% for testing using stratified sampling. Class distribution in the training set includes 34.7\% sexist texts and 65.3\% non-sexist texts. We perform cross validation experiments on the training data to fine-tune model hyperparameters, choosing the best models for the final experiments. We report global macro F1 and accuracy scores for the three tasks, as well as F1 scores specific to each class for experimental step 1 and weighted F1 scores for steps 2 and 3.

\subsection{Experiment Results and Analysis}

\begin{table*}[htb]
\centering
\caption{Sexism detection performance. F1-Sex and F1-Not denote F1 scores respectively for binary labels of sexist or non-sexist. mF1 denotes macro F1 score and Acc denotes accuracy score.}
\label{tab:result}
\resizebox{0.99\textwidth}{!}{%
\begin{tabular}{lcccccccc}
\hline
\textbf{Model} & \multicolumn{4}{c}{\textbf{Original Feature}}                       & \multicolumn{4}{c}{\textbf{+Lexicon}}                               \\ \cline{2-9} 
                                & \textbf{F1-Sex} & \textbf{F1-Not} & \textbf{mF1}   & \textbf{Acc}   & \textbf{F1-Sex} & \textbf{F1-Not} & \textbf{mF1}   & \textbf{Acc}   \\ \hline
LR + ngram                      & 0.624           & 0.849           & 0.737          & 0.785          & 0.616           & 0.846           & 0.731          & 0.780          \\ \hline
Char-LR + ngram                 & 0.640           & 0.852           & 0.746          & 0.790          & 0.646           & \textbf{0.858}  & 0.752          & 0.797          \\ \hline
SVM + ngram                     & 0.633           & 0.844           & 0.739          & 0.781          & 0.640           & 0.842           & 0.741          & 0.786          \\ \hline
CNN + ft                        & 0.669           & 0.828           & 0.749          & 0.774          & 0.654           & 0.844           & 0.749          & 0.785          \\ \hline
Char-CNN + ft                   & 0.660           & 0.845           & 0.753          & 0.787          & 0.654           & 0.850           & 0.752          & 0.790          \\ \hline
Bert                            & \textbf{0.694}  & \textbf{0.858}  & \textbf{0.776} & \textbf{0.806} & 0.661           & 0.844           & 0.752          & 0.786          \\ \hline
Bert-wwm                        & 0.678           & 0.846           & 0.762          & 0.792          & 0.699           & 0.851           & 0.775          & 0.800          \\ \hline
RoBerta                         & 0.685           & 0.844           & 0.764          & 0.792          & \textbf{0.707}  & 0.853           & \textbf{0.780} & \textbf{0.804} \\ \hline
\end{tabular}%
}
\end{table*}

From the results in Table \ref{tab:result}, we see that content-based models (CNN) outperform linguistic ones (LR / SVM) in both word level and character level while context-based models (BERT) perform best. Character-level models (e.g. char-LR / char-CNN) show better performance than word-level models (e.g. LR / CNN), proving them more suitable for a language like Chinese with no space between words. When we incorporate lexical features, most models lead to slight improvements of 0.5-1\% in F1 score (with the exception of LR and BERT models), showing the potential of SexHateLex in improving performance, particularly with the best-performing model RoBERTa. We also observe an overall tendency for achieving 15-23\% better prediction on the non-sexist category, highlighting the challenge of detecting sexist comments.

\begin{table*}[]
\centering
\caption{Results for the sexism category and target classification tasks. mF1 denotes macro F1 score and wF1 denotes weighted F1 score. Acc denotes accuracy score.}
\label{tab:task23}
\resizebox{0.77\textwidth}{!}{%
\begin{tabular}{lcccccc}
\hline
\textbf{Model} & \multicolumn{3}{c}{\textbf{Category classification}}               & \multicolumn{3}{c}{\textbf{Target classification}} \\ \cline{2-7} 
                                & \multicolumn{1}{l}{\textbf{wF1}} & \textbf{mF1}   & \textbf{Acc}   & \textbf{wF1}    & \textbf{mF1}    & \textbf{Acc}   \\ \hline
LR + ngram                      & 0.628                            & 0.310          & 0.611          & 0.663           & 0.447           & 0.719          \\ \hline
Char-LR + ngram                 & 0.648                            & 0.316          & 0.646          & 0.657           & 0.428           & 0.721          \\ \hline
SVM + ngram                     & 0.647                            & 0.320          & 0.692          & 0.661           & 0.446           & 0.707          \\ \hline
CNN + ft                        & 0.711                            & 0.335          & 0.716          & 0.668           & 0.447           & 0.711          \\ \hline
Char-CNN + ft                   & 0.722                            & 0.347          & 0.730          & 0.670           & 0.448           & 0.714          \\ \hline
Bert                            & 0.732                            & 0.355          & \textbf{0.736} & 0.678           & 0.457           & 0.713          \\ \hline
Bert-wwm                        & 0.732                            & 0.354          & \textbf{0.736} & 0.682           & 0.462           & 0.720          \\ \hline
RoBerta                         & \textbf{0.734}                   & \textbf{0.360} & 0.732          & \textbf{0.687}  & \textbf{0.467}  & \textbf{0.727} \\ \hline
\end{tabular}%
}
\end{table*}

Regarding the category classification task, the results in Table \ref{tab:task23} show a different scenario. The best performing model is RoBERTa, with highest weighted and F1 scores, but all three BERT-based models have better performance than others. For the third task, the results in Table \ref{tab:task23} show that all the models achieve a competitive performance without a large margin, while RoBERTa performs best across other models. Besides, it can be observed that macro F1 scores for both task 2 and 3 show an averaged lower than weighted F1 scores, which indicates a potential impact of the imbalanced nature of the data among the finer-grained classes. More sampling methods are supposed to be considered before training.

\section{Discussion}

\subsection{Error Analysis}

\begin{table*}[htb]
\centering
\caption{Error analysis for misclassified examples. TL denotes true label and PL denotes predicted label.}
\label{tab:error}
\resizebox{0.99\textwidth}{!}{%
\begin{tabular}{lllll}
\hline
\textbf{Error Type} & \textbf{Example}                                                 & \textbf{Translation}                                                                                                                 & \textbf{TL} & \textbf{PL} \\ \hline
(1)         & \begin{tabular}[c]{@{}l@{}}如果她自己够优秀就不会\\ 在网络上怨天尤人了\end{tabular}  & \begin{tabular}[c]{@{}l@{}}If she is excellent enough, she won't blame \\ others on the Internet\end{tabular}                         & 1                   & 0                        \\ \hline
(2)         & \begin{tabular}[c]{@{}l@{}}你这种金针菇明码标价了\\ 也只会烂在货架上\end{tabular}   & \begin{tabular}[c]{@{}l@{}}Enoki mushrooms like yours will only rot \\ on the shelf even if they are clearly marked\end{tabular} & 1                   & 0                        \\ \hline
(3)         & \begin{tabular}[c]{@{}l@{}}田园女权，女拳师，极端\\ 女权，是我是我都是我\end{tabular} & \begin{tabular}[c]{@{}l@{}}Pastoral feminist, female boxer, extreme \\ feminist, it’s all me\end{tabular}                        & 0                   & 1                        \\ \hline
\end{tabular}%
}
\end{table*}

We look at frequent errors across misclassified instances generated from SVM, CNN and BERT, three typical models selected from three types of models we used in the experiment step 1 (see Table \ref{tab:error} for examples). Several typical errors appeared in the experiments are summarised below:

\textbf{(a) Implicit sexism: } Errors in those posts lacking explicit sexist expression or context, and most frequent reason for (a) in the misclassified texts is caused by sarcastic expressions. Sarcasm seems to be a suitable way for expressing contempt and subtly offending individuals, which modifies the perception of message, hindering the correct detection of sexism by automatic systems \cite{Thomae2015sexist}. Example (1) is a sarcastic comment that criticises women who are not successful but insults those people who uphold gender equality. It is difficult to identify sexism when there is no explicit presence of abusive language. Another problem is that the model cannot pick up words with a specific meaning related to gender.

\textbf{(b) Lack of prior information: } It demonstrates that the model cannot identify those contents referring to sexism-related event, people or words/phrases with special meanings as it does not possess prior knowledge. In example (2), 金针菇 (enoki mushroom) is a very harmful word specifically towards men associated with some physical characteristics but cannot be directly identified by the model. 

\textbf{(c) Overuse of sexist words: } It indicates that sexist words might be overused in one text, leading to the over-dependence of the model on these words, while sexist targets in posts are confounding and hard to be identified. We can see from example (3) that the model can easily identify a text with many sexist words as a sexist text even if there is no specific targeted individual or group attacked by someone. 


\subsection{External Knowledge Induction}

After infusing external domain information to models, most of them present a slight increase in the final performance. We conjecture on a set of factors that may be affecting the performance of using the lexicon:

\begin{itemize}
    \item \textbf{Dataset variety:} The lexical terms found in the randomly split training and test sets might be imbalanced. There may be a certain gap in the quantity of lexical terms extracted in the proportion of the training and test sets, leading to the diverse degree of the influence of lexical terms in the process of the model classification.
    \item \textbf{Term inconsistency between dataset and lexicon:} Terms in the dataset and the lexicon could be inconsistent. The domain-specific lexicon might not be capable of covering all sexism-related terms encountered across datasets.
    \item \textbf{Linguistic characteristics:} Not all posts containing hateful terms are sexist necessarily, due to cases of polysemy or negation.
    \item \textbf{Humour, irony and sarcasm:} Sexist posts with humour, irony and sarcasm are implicit and difficult to be identified, and may contain no explicit hate-related terms.
    \item \textbf{Spelling variation:} Spelling variation is prevalent in social media \cite{vidgen-etal-2019-challenges}. Sensitive words sometimes use spelling variations to obfuscate and avoid detection, which do not match those normative words in the lexicon. A certain Chinese character in a word (e.g.绿茶婊$\rightarrow$angelic bitch\footnote{绿茶婊(angelic bitch) means girls who pretend to be pure and innocent but in fact are manipulative and scheming.}) is often replaced by homophones or pinyin (e.g. 绿茶婊$\rightarrow$绿茶表/绿茶biao), or is split into radicals according to the composition rules of Chinese characters (e.g. 绿茶婊$\rightarrow$绿茶女表).
    \item \textbf{Quality of lexical features:} TF-IDF frequency features captured from the category of lexical terms might be comparatively sparse and lose information for specific terms. Lexical embeddings derived from pre-trained word embedding models could be beneficial as high-quality word embeddings can be learned efficiently thanks to low space and time complexity \cite{goldberg2014word2vec}.
    \item \textbf{Approaches for lexicon induction:} Since the approach for lexicon induction might not fully absorb lexical information by simple concatenation between textual hidden features and lexical features, other forms of fusion can be tested, such as matrix multiplication \cite{pappas2019gile} and cosine similarity \cite{li2020label}.
\end{itemize}

\section{Research Applications}
\label{sec:application}

The SWSR dataset and the SexHateLex lexicon provide resources for furthering research in a new language in the growing research problem of sexist language. We discuss potential areas of research.

\subsection{User-based Sexism Detection}

As sexism-related speech belonging to user-generated content online, some investigations are conducted to find out the potential influence of user characteristics like gender and location on sexism detection \cite{waseem2016hateful}. User metadata in SWSR, such as gender, location and number of followings, can enable researchers to explore possible correlations between gender-based hateful content and user profiles, furthering user-based studies in the area of sexism detection.

\subsection{Explainable Sexism Detection}

Providing explanations can make model outputs more convincing and understandable \cite{molnar2020interpretable,dai2020ginger}. We provide our dataset with two basic classes to show which text is sexist or not, with fine-grained labels to support furthering detection. Besides, we offer a lexicon composed of abusive words to support detection of offensive content with sexism-specific features.

\subsection{Multi-lingual and Cross-lingual Sexism Detection}

While most approaches to sexism detection have been proposed for English, other studies have been investigated to deal with this task in other languages such as Spanish, Italian, and Indian, thanks to recent shared tasks \cite{fersini2018ibereval,fersini2018evalita,mandl2019overview}. More research is needed in other languages, including Chinese, both in multilingual settings, i.e. proposing models that deal with multiple languages, and cross-lingual settings, i.e. leveraging data in a resource-rich language like English for application in lesser-resourced languages such as Chinese. Our dataset compensates for the lack of sexist speech in Chinese, thereby facilitating the development of sexism identification research in multi-lingual and cross-lingual settings.

\subsection{Cross-domain Hate Speech Detection}

With the prevalence of identifying hate speech online, some studies concentrate on detecting specific types of hate speech, such as racism or sexism. This differences across types of hate speech make it more challenging to generalise hate speech detection models. Cross-domain detection of hate speech thereby has been a topic of interest to identify common features between distinct hate speech domains, achieving knowledge transfer and model generalization. Our dataset provides gender-related hateful texts with corresponding topic-related keywords, which could enhance research on sexism and facilitate potential research of cross-domain detection in this and other types of hate speech, particularly if additional Chinese hate speech datasets are released.

\subsection{Other applications}

While most existing research on sexism detection focus on detecting the text to binary classes (sexist or not), our dataset enables investigation of additional, finer-grained perspectives of sexism, thanks to three types of labels provided. Categorising sexism by type as well as identifying the type of targets enable furthering research in sexism detection beyond the widely-studied binary classification task.

\section{FAIR}
\label{sec:fair}

In this section, we show that the SWSR dataset is collected and organised based on 'FAIR' facets: Findable, Accessible, Interoperable and Re-usable. Our dataset is publicly available through Zenodo and can be downloaded completely by using the following citation:

Aiqi Jiang, Xiaohan Yang, Yang Liu, \& Arkaitz Zubiaga. (2021). SWSR: A Chinese Dataset and Lexicon for Online Sexism Detection [Dataset]. Zenodo. http://doi.org/10.5281/zenodo.4773875

The dataset files are provided in CSV format for the SWSR dataset and TXT format for the SexHateLex lexicon. A README file is included to explain each file in detail to facilitate the re-use of the dataset.

\section{Conclusion}
\label{sec:conclusion}

In this paper, we release a comprehensive sexism dataset SWSR along with a large lexicon SexHateLex, to facilitate research on online gender-based speech in Chinese. To the best of our knowledge, this is the first sexism dataset in Chinese. The dataset provides both weibo and comment texts, as well as three types of labels, namely sexist or not, sexism category and target type. The dataset contains two files for \textit{SexComment} and \textit{SexWeibo}, containing sexist comments, original weibos enabling contextual analysis, and anonymised user metadata. We further conduct exploratory analyses of the dataset. Different types of sexism detection approaches are also evaluated on \textit{SexComment}. We experiment with baseline models for sexism detection, which provides a benchmark for further experimentation. We expect our dataset to enable further research in Chinese sexism detection, including a set of possible directions.

\section{Acknowledgements}

Aiqi Jiang is funded by China Scholarship Council (CSC). This research utilised Queen Mary’s Apocrita HPC facility, supported by QMUL Research-IT http://doi.org/10.5281/zenodo.438045. Besides, we are grateful to Peiling Yi, Xia Zeng, Xiaoyu Guo and Wenjie Yin for their assistance discussing the annotation guidelines.


\bibliography{mybibfile.re}

\appendix

\section{SWSR Dataset Format}
\label{sec:appendix-structure}

SWSR dataset consists of two files: 'SexWeibo.csv' and 'SexComment.csv', containing weibos (posts) and comments (replies) respectively. See more detailed description of features below:

\subsection{SexWeibo.csv}

\begin{itemize}
    \item weibo\_id: a string of weibo ID 
    \item weibo\_text: a string of weibo content
    \item keyword: contains sexism-related keyword(s) extracted from the weibo text
    \item user\_gender: the gender of user
    \item user\_location: the location of user
    \item user\_follower: number of users who follow this user's account
    \item user\_following: number of users whom this user follows
    \item weibo\_like: number of like for the weibo
    \item weibo\_comment: number of comment for the weibo
    \item weibo\_repost: number of repost for the weibo
    \item weibo\_date: the date and time when the weibo is posted
      
\end{itemize}

\subsection{SexComment.csv}

\begin{itemize}
    \item weibo\_id: the weibo id where the comment is collected
    \item comment\_text: a string of the comment
    \item gender: the gender of commenter
    \item location: the location of commenter
    \item like: number of like for this comment
    \item date: the date and time when the comment is posted
    \item label: the comment is sexist(1) or non-sexist(0)
    \item category: categorise sexism into four classes -- Stereotype based on Appearance(SA), Stereotype based on Cultural Background (SCB), MicroAggression (MA) and Sexual Offense (SO)
    \item target:  the type of target who are attacked -- Individual (I) or Generic (G)
\end{itemize}

\end{CJK*}
\end{document}